\documentclass[runningheads,a4paper]{llncs}

\usepackage{amssymb}
\usepackage{graphicx}

\usepackage{amsmath}
\usepackage{amssymb}
\usepackage{latexsym}
\usepackage{url}
\usepackage{cite}
\usepackage{relsize}
\usepackage{multirow}
\usepackage{afterpage}
\usepackage{ifthen}
\usepackage{graphicx}
\usepackage{algpseudocode,algorithm,algorithmicx}
\usepackage{subfigure}
\usepackage[keeplastbox]{flushend}
\usepackage{epstopdf}
\usepackage{stackengine}
\usepackage{gensymb}
\usepackage[bookmarks=false, linkcolor=blue, urlcolor=blue, citecolor=blue]{hyperref} 
\usepackage{booktabs}
\usepackage{makecell}
\usepackage{hhline}
\usepackage{courier}
\usepackage{lipsum}
\usepackage{xcolor}
\usepackage{siunitx}

\hypersetup{pdfstartview={FitH}}

\usepackage{listings}

\usepackage{url}
\urldef{\mailsa}\path|a.nguyen@imperial.ac.uk|
\urldef{\mailsb}\path|quang.tran@aioz.io|
\urldef{\mailsc}\path||    
\newcommand{\keywords}[1]{\par\addvspace\baselineskip
\noindent\keywordname\enspace\ignorespaces#1}

\newcommand{\ra}[1]{\renewcommand{\arraystretch}{#1}}

\begin{document}

\mainmatter  

\title{\LARGE Autonomous Navigation with Mobile Robots using Deep Learning and the Robot Operating System
}

\titlerunning{Autonomous Navigation with Deep Learning}

%
%
\author{Anh Nguyen$^{1, *}$,
        Quang D. Tran$^{2}$
}

\authorrunning{Anh Nguyen et al.}


\institute{$^{1}$ Department of Computing, Imperial College London, UK \\
            \email{a.nguyen@imperial.ac.uk} \\
            $^{2}$ AIOZ Ltd., Singapore \\ \email{quang.tran@aioz.io} \\
            $^{*}$ Corresponding author
}

%
%

\toctitle{Autonomous Navigation with Deep Learning}
\tocauthor{Anh Nguyen et al.}
\maketitle

\begin{abstract}
Autonomous navigation is a long-standing field of robotics research, which provides an essential capability for mobile robots to execute a series of tasks on the same environments performed by human everyday. In this chapter, we present a set of algorithms to train and deploy deep networks for autonomous navigation of mobile robots using the Robot Operation System (ROS). We describe three main steps to tackle this problem: \textit{(i)} collecting data in simulation environments using ROS and Gazebo; \textit{(ii)} designing deep network for autonomous navigation, and \textit{(iii)} deploying the learned policy on mobile robots in both simulation and real-world. Theoretically, we present deep learning architectures for robust navigation in normal environments (e.g., man-made houses, roads) and complex environments (e.g., collapsed cities, or natural caves). We further show that the use of visual modalities such as RGB, Lidar, and point cloud is essential to improve the autonomy of mobile robots. Our project website and demonstration video can be found at \url{https://sites.google.com/site/autonomousnavigationros}
\keywords{Mobile robot, Deep learning, Autonomous navigation}
\end{abstract}

\section{Introduction}

Autonomous navigation has a long history in robotics research and attracts a lot of work in both academia and industry. In general, the task of autonomous navigation is to control a robot navigate around the environment without colliding with obstacles. It can be seen that navigation is an elementary skill for intelligent agents, which requires decision-making across a diverse range of scales in time and space. In practice, autonomous navigation is not a trivial task since the robot needs to sense the environment in real-time and react accordingly. The problem becomes even more difficult in real-world settings as the sensing and reaction loop always have noise or uncertainty (e.g., imperfect data from sensors, or inaccuracy feedback from the robot's actuator). 

\begin{figure}[!t] 
    \centering
    \includegraphics[width=0.99\linewidth, height=0.55\linewidth]{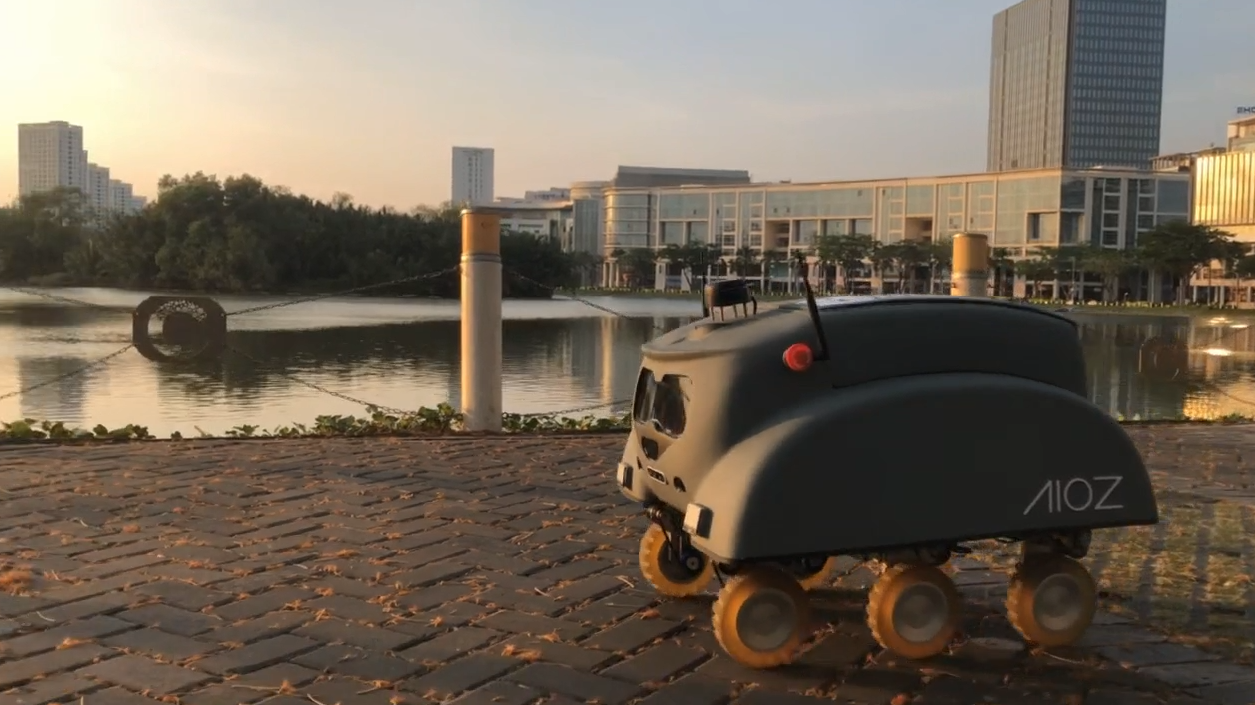} 
    \vspace{3ex}
    \caption{BeetleBot is navigating through its environment using a learned policy from the deep network and the Robot Operating System.}
    \label{Fig:intro_robot} 
\end{figure}

 With the rise of deep learning, learning-based methods have become a popular approach to directly derive end-to-end policies which map raw sensor data to control commands~\cite{pfeiffer2017perception, nguyen2019v2cnet}. This end-to-end learning effectively utilizes input data from different sensors (e.g., depth camera, IMU, laser sensor) thereby reducing  power consumption and processing time. Another advantage of this approach is the end-to-end relationship between data input (i.e. sensor) and control outputs (i.e. actuator) can result in an arbitrarily non-linear complex function, which has showed encouraging results in different navigation problems such as autonomous driving~\cite{BojarskiTDFFGJM16} or UAV control~\cite{Monajjemi16}. In this chapter, we present the deep learning models to allow a mobile robot to navigate in both man-made and complex environments such as collapsed houses/cities that suffered from a disaster (e.g. an earthquake) or a natural cave (Fig.~\ref{Fig:intro_robot}). We further provide the practical experiences to collect training data using Gazebo and deploy deep learning models to mobile robots using using the Robot Operating System (ROS).

\begin{figure}[h] 
    \centering
    \includegraphics[width=0.99\linewidth, height=0.6\linewidth]{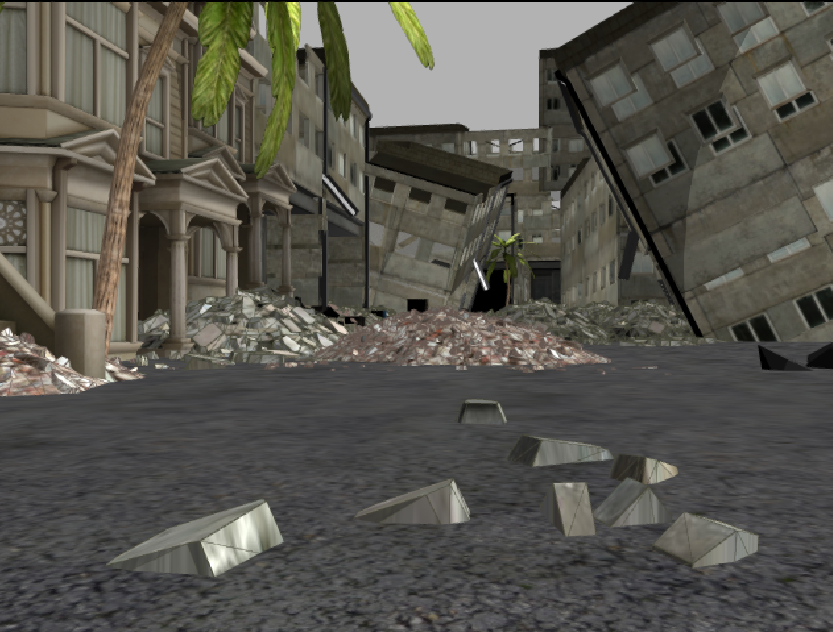} 
    \vspace{3ex}
    \caption{An example of a collapsed city in Gazebo simulation.}
    \label{Fig:intro_city} 
\end{figure}

While the normal environments usually have clear visual information in normal condition, complex environments such as collapsed cities or natural caves pose significant challenges for autonomous navigation~\cite{piotr_2017_navigate_in_complex_environments}. This is because the complex environments usually have very challenging visual or physical properties. For example, the collapsed cities may have constrained narrow passages, vertical shafts, unstable pathways with debris and broken objects; or the natural caves often have irregular geological structures, narrow passages, and poor lighting condition. Autonomous navigation with intelligent robots in complex environments, however, is a crucial task, especially in time-sensitive scenarios such as disaster response, search and rescue, etc. Fig \ref{Fig:intro_city} shows an example of a collapsed city built on Gazebo that is used to collect data to train a deep navigation model in our work.

\section{Related Work} \label{Sec:rw}
Autonomous navigation is a popular research topic in robotics~\cite{kam1997sensor}. Traditional methods tackle this problem using algorithms based on Kalman Filter~\cite{dobrev2016multi} for sensor fusion. In~\cite{lynen2013robust}, the authors proposed a method based on Extended Kalman Filter (EKF) for AUV navigation. Similarly, the work in~\cite{du2020real} developed an algorithm based on EKF to estimate the state of an UAV in real-time. Apart from the traditional localization and navigation task, multimodal fusion is also used in other applications such as visual segmentation~\cite{valada2017adapnet} or object detection or captioning~\cite{mees16iros, nguyen2019object, do2018affordancenet} in challenging environments. In both~\cite{mees16iros, valada2017adapnet} multimodal inputs are combined from different sensors to learn a deep policy in challenging lighting environments.

With the recent advancement in machine learning, many works have been proposed to directly learn control policies from raw sensory data. These methods can be grouped into two categories: reinforcement learning methods~\cite{duan2016benchmarking} and supervised learning methods~\cite{ZhangC16b, XuGYD16}. In~\cite{bojarski2016end2end_car}, the authors proposed the first end-to-end navigation system for autonomous car using 2D images. Smolyanskiy et al.~\cite{smolyanskiy2017toward} applied this idea on UAV using data from three input cameras. Similarly, DroNet~\cite{loquercio2018dronet} used CNN to learn the steering angle and predict the collision probability given the RGB input image. Gandhi et al.~\cite{gandhi2017learning} introduced a navigation method for UAV by learning from negative and positive crashing trials. Monajjemi et al.~\cite{Monajjemi16} proposed a new method for agile UAV control. The work of~\cite{amini2018variational} combined CNN and Variational Encoder to estimate the steering control signal. The authors in~\cite{alexander_2019_variational_end2end} combined the navigation map with visual input to learn the control signal for autonomous navigation.

Apart from CNN-based methods, reinforcement learning algorithms are also widely used to learn control policies from robot experiences~\cite{duan2016benchmarking, schulman2015trust}. In~\cite{zhu2017target, delbrouck2018_object_navigation}, the authors addressed the target-driven navigation problem given an input picture of a target object. Wortsman et al.\cite{wortsman2018_RL} developed a self-adaptive visual navigation system using reinforcement learning and meta-learning. The authors in~\cite{sadeghi2016cad2rl}\cite{mancini2017toward} trained the reinforcement policy agents in simulation environments, then transfer the learned policy to the real-world. In~\cite{andersson2017deep}~\cite{dosovitskiy2016learning}, the authors combined deep reinforcement learning with CNN to leverage the advantages of both techniques. The authors in~\cite{piotr_2017_navigate_in_complex_environments} developed a method to train autonomous agents to navigate within large and complicated environments such as 3D mazes.

In this chapter, we choose the end-to-end supervised learning approach for the ease of deploying and testing in real robot systems. We first simulate the environments in physics-based simulation engine and collect a large-scale for supervised learning. We then design and trained the network to learn the control policy. Finally, the learned policy is deployed on mobile robots in both simulation and real-world scenarios.   
\section{Data Collection}
To train a deep network, it is essential to collect a large-scale dataset with ground-truth for supervised learning. Similar to our previous work~\cite{nguyen2020autonomous}, we create the simulation models of these environments in Gazebo and ROS to collect the visual data from simulation. In particular, we collect the data from these types of environment: 
\begin{itemize}
\item Normal city: A normal city environment with man-made road, building, tree, road, etc.
\item Collapsed house: The house or indoor environment that suffered from an accident or a disaster (e.g. an earthquake) with random objects on the ground.
\item Collapsed city: The outdoor environment with many debris from the collapsed house/wall. 
\item Natural cave: A long tunnel with poor lighting condition and irregular geological structures.
\end{itemize}

To build the simulation environments, we first create the 3D model of normal daily objects in indoor and outdoor environments (e.g. beds, tables, lamps, computers, tools, trees, cars, rocks, etc.), including broken objects (e.g. broken vases, broken dishes, and debris). These objects are then manually chosen and placed in each environment to create the entire simulated environment. The world file of these environment is Gazebo and ROS friendly and can be downloaded from our project website.

For each environment, we use a mobile robot model equipped with a laser sensor and a depth camera mounted on top of the robot to collect the visual data. The robot is controlled manually to navigate around each environment. We collect the visual data when the robot is moving. All the visual data are synchronized with a current steering signal of the robot at each timestamp. 

\begin{figure*}[!t] 
    \centering
    \includegraphics[width=0.99\linewidth, height=0.75\linewidth]{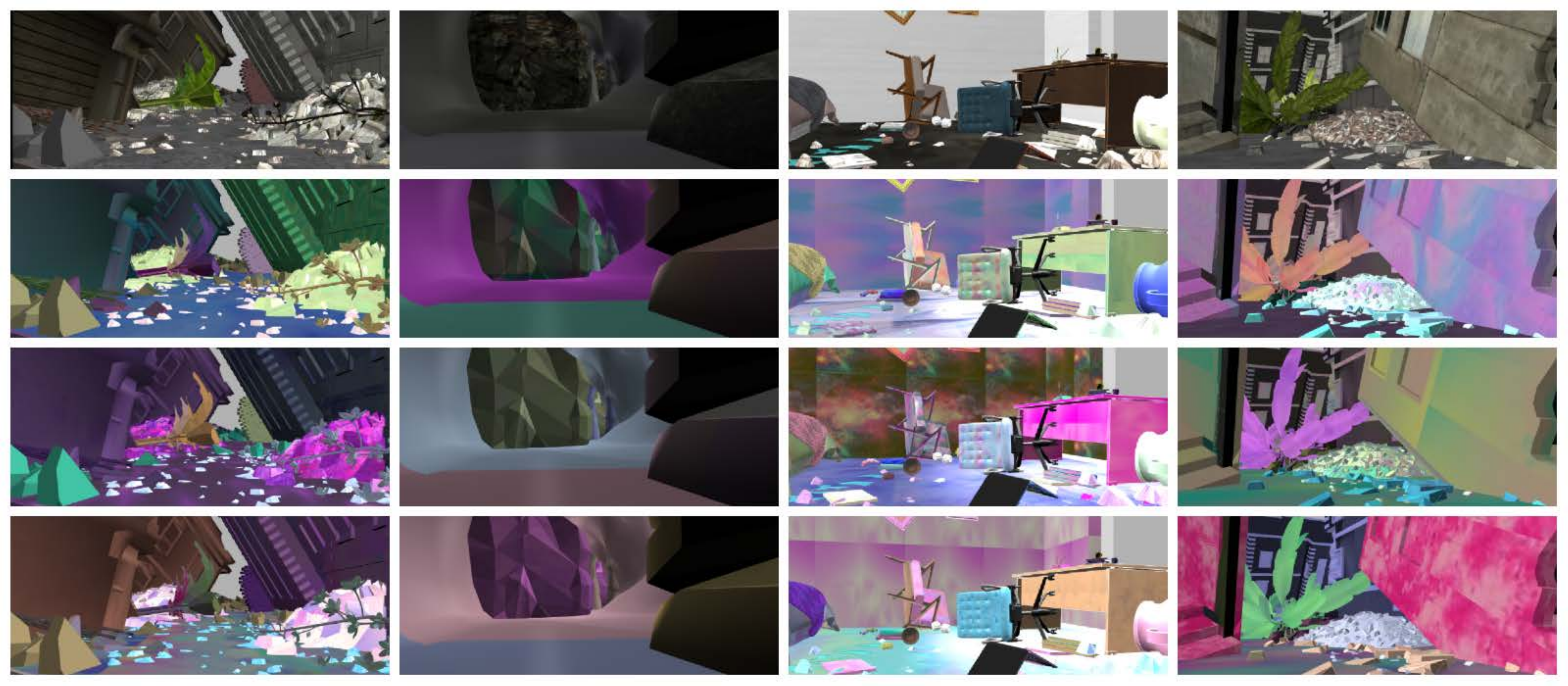} 
    \vspace{3ex}
    \caption{Different robot's views in our simulation of complex environments: collapsed city, natural cave, and collapsed house. \textbf{Top row:} RGB images from robot viewpoint in normal setting. \textbf{Other rows:} The same viewpoint when applying domain randomisation to the simulated environments.}
    \label{Fig_dataset} 
\end{figure*}

\textbf{Data Statistic}
In particular, we create $539$ 3D object models to build both normal and complex environments. In average, the collapsed house environments are built with approximately $130$ objects in an area of $400m^2$. The collapsed city and man-made road has $275$ objects and spread in $3,000m^2$, while the natural cave environments are built with $60$ objects in approximately $4,000m^2$ area. We manually control a mobile robot in $40$ hours to collect the data. 

In total, we collect around $40,000$ visual data triples (RGB image, point cloud, distance map) for each environment type, resulting a large-scale dataset with $120,000$ records of synchronized RGB image, point cloud, laser distance map, and ground-truth steering angle. Around $45\%$ of the dataset are collected when we use domain randomisation by applying random texture to the environments (Fig.~\ref{Fig_dataset}). For each environment, we use $70\%$ data for training and $30\%$ data for testing.  All the 3D environments and our dataset are publicly available and can be found in our project website.

Apart from training with normal data, we also employ the training method using domain randomisation~\cite{james2017transferring}. As shown in~\cite{james2017transferring}, this simple technique can effectively improve the generalization of the network when only simulation data are available for training.
\section{Network Architecture} \label{Sec_visual}
In this section, we first present a deep model for autonomous navigation in man-made road using only RGB images as the input. We then present a more advance model to navigate in complex environments using sensor fusion.

\begin{figure*}[h] 
    \centering
    \includegraphics[width=0.99\linewidth, height=0.4\linewidth]{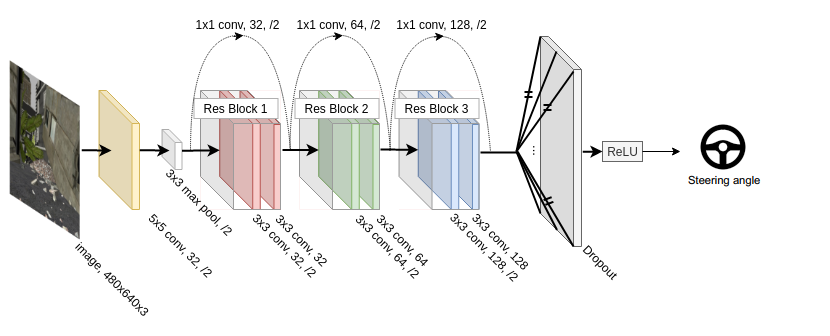} 
    \vspace{3ex}
    \caption{A deep network architecture using ResNet8 to predict the steering angle for autonomous navigation. }
    \label{Fig_resnet8_arc} 
\end{figure*}

\begin{figure*}[h] 
    \centering
    \includegraphics[width=0.7\linewidth, height=0.2\linewidth]{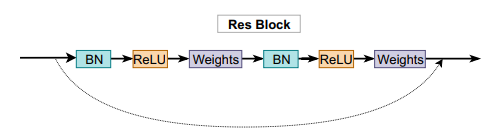} 
    \vspace{3ex}
    \caption{A detailed visualization of a ResNet8's block.}
    \label{Fig_resnet_block} 
\end{figure*}

\subsection{Navigation in Man-made Road} \label{nav_man_made}
Navigation in man-made road is one of the hot research topic recently in robotic. With many applications such as autonomous car, this topic attracts interests from both academia and industry. With some assumptions such as there are no dynamic obstacles and the environment is well-controlled, we can build and deploy and deep network to address this problem in simulation. Fig. \ref{Fig_resnet8_arc} shows an overview of our approach.

As in all other visual recognition tasks, learning meaningful features from 2D images is the key to success in our autonomous navigation system. In this work, we use ResNet8~\cite{he2016deep} to extract deep features from the input RGB images. The ResNet8 has $3$ residual blocks, each block consists of a convolutional layer, ReLU, skip links and batch normalization operations. A  residual block is skip-connection block that learn residual functions of the input layers, instead of learning unreferenced functions. The intuition is that it is easier to optimize the residual loss than to optimize the unreferenced mapping. With the skip connections of the residual block, the network can be deeper while being more robust against the vanishing gradient problem. A detailed visualization of ResNet8's block can be found in Fig.~\ref{Fig_resnet_block}.

As in~\cite{loquercio2018dronet}, we use ResNet8 to learn deep features from the input 2D images since it is a light weight network, achieving competitive performance, and easy to deploy on physical robots. At the end of the network, we use a Dropout layer to prevent the overfitting during the training. The network is trained end-to-end to regress a steering value that present the action of the current control state of the robot. With this architecture, we can have a deep network to learn and deploy in simple man-made scenarios, however, this architecture does not generalize well on more complicated scenarios such as dealing with dynamic obstacles or navigating through complex environments.

\subsection{Navigation in Complex Environment}
Complex environments such as natural cave networks or collapsed cities pose significant challenges for autonomous navigation due to their irregular structures, unexpected obstacles, and the poor lighting condition inside the environments. To overcome these natural difficulties, we use three visual input data in our method: RGB image $\mathcal{I}$, point cloud $\mathcal{P}$, and distance map $\mathcal{D}$ obtaining from the laser sensor. Intuitively, the use of all three visual modalities ensures that the robot's perception system has meaningful information from at least one modality during the navigation under challenging conditions such as lighting changes, sensor noise in depth channels due to reflective materials, or motion blur, etc. 

In practice, the RGB images and point clouds are captured using a depth camera mounted in front of the robot while the distance map is reconstructed from the laser data. In complex environments, while the RGB images and point clouds can provide the visual semantic information for the robot, the robot may need more useful information such as the distance map due to the presence of various obstacles. The distance map is reconstructed from the laser data as follows:


\begin{equation}
\begin{aligned}
x_i &= x_0 + d*cos(\pi - \phi * i) \\
y_i &= y_0 - d*sin(\phi * i)
\end{aligned}
\end{equation}
where $x_i,y_i$ is the coordinate of $i^{th}$ point on 2D distance map. $x_0,y_0$ is the coordinate of robot. $d$ is the distance from the laser sensor to the obstacle, and $\phi$ is the incremental angle of the laser sensor.

\begin{figure}[!ht] 
    \centering
    \includegraphics[width=0.99\linewidth, height=0.75\linewidth]{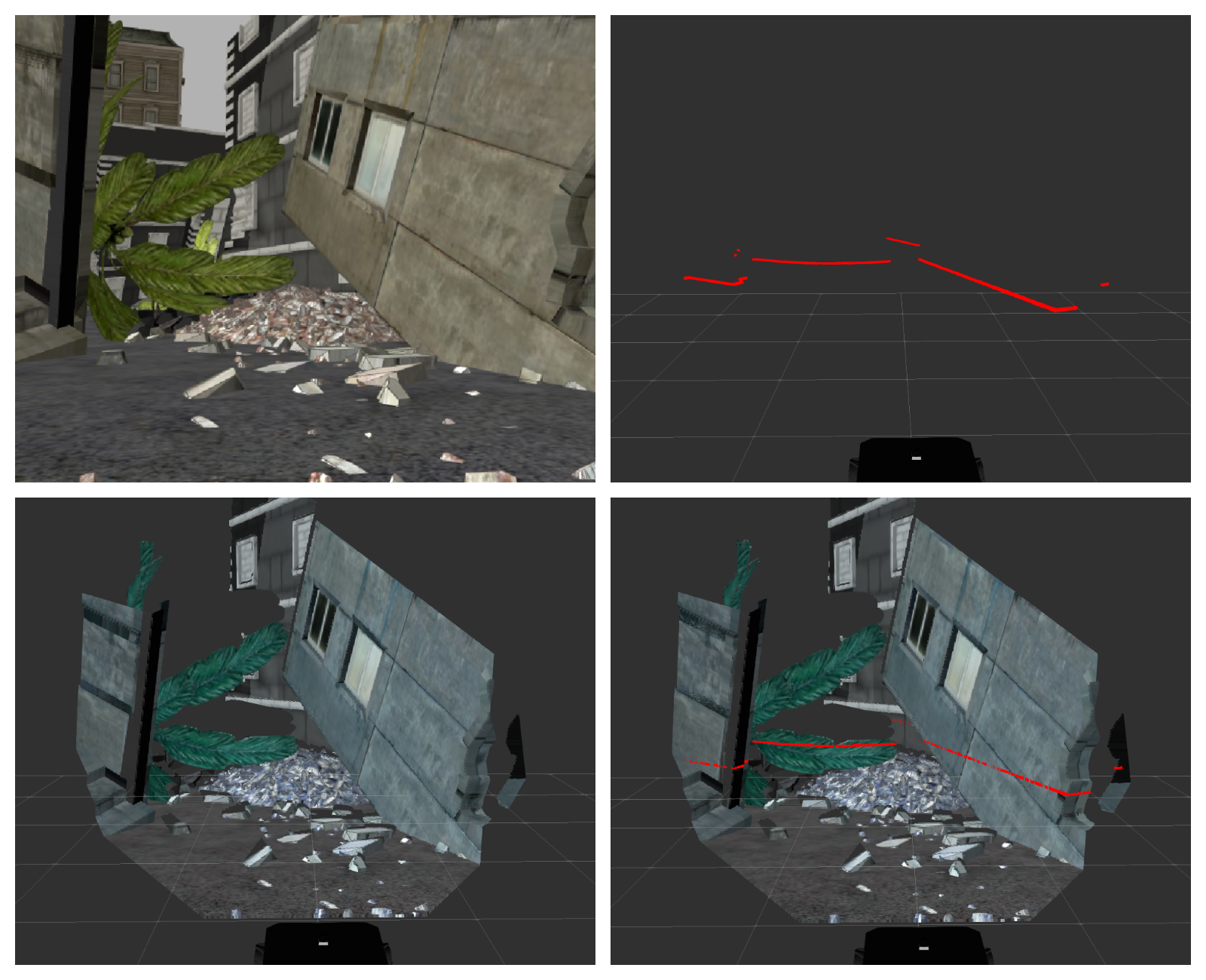} 
    \vspace{2ex}
    \caption{An illustration of three visual modalities used in our network. \textbf{Top row:} The RGB image (left) and the distance map from laser scanner (right). \textbf{Bottom row:} The 3D point cloud (left) and the overlay of the distance map in 3D point cloud (right).}
    \label{Fig_visual_input} 
\end{figure}

To keep the low latency between three visual modalities, we use only one laser scan to reconstruct the distance map. The scanning angle of the laser sensor is set to $180^\circ$ to cover the front view of the robot, and the maximum distance is approximately $16m$. This will help the robots aware of the obstacles from its far left/right hand side, since these obstacles may not be captured in the front camera which provides the RGB images and point cloud data. We notice that all three modalities are synchronized at each timestamp to ensure the robot is observing the same viewpoint at each control state. Fig.~\ref{Fig_visual_input} shows a visualization of three visual modalities used in our method.

\begin{figure*}[!ht] 
    \centering
    \includegraphics[width=0.99\linewidth, height=0.5\linewidth]{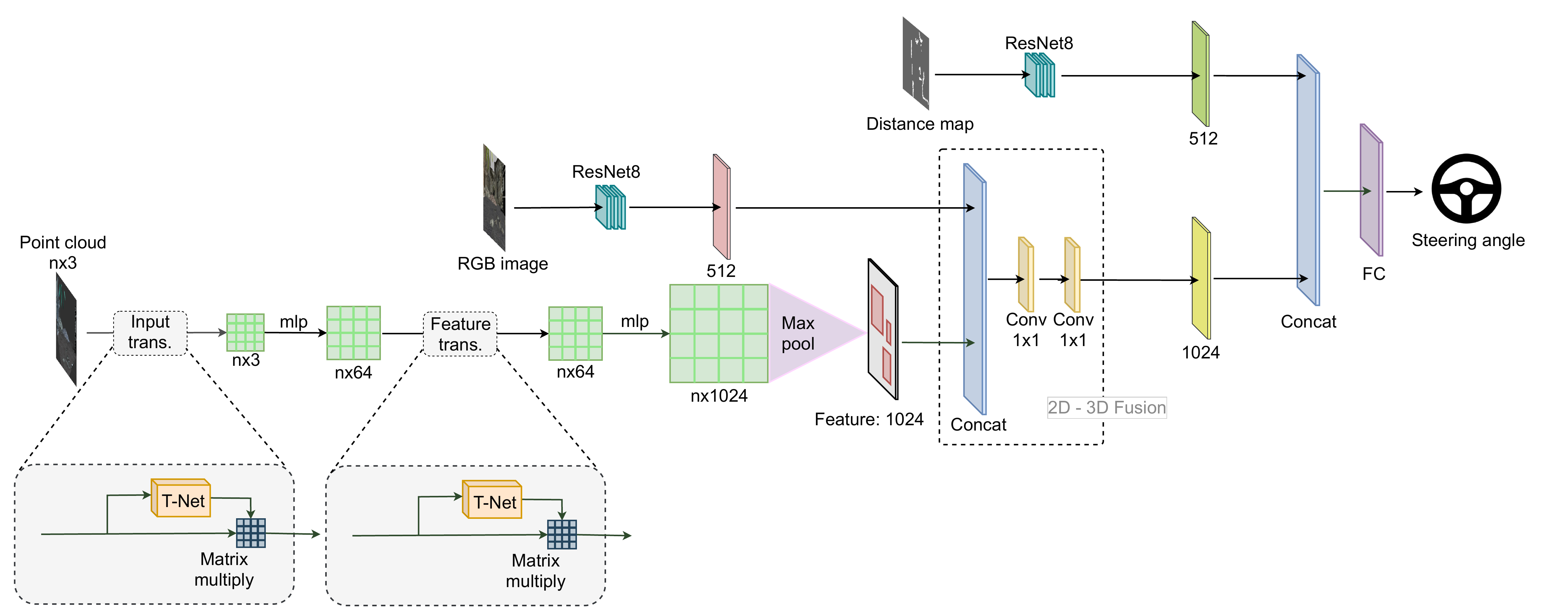} 
    \vspace{4ex}
    \caption{An overview of our NMFNet architecture. The network consists of three branches: The first branch learns the depth features from the distance map. The second branch extracts the features from RGB images, and the third branch handles the point cloud data. We then employ the 2D-3D fusion to predict the steering angle.}
    \label{Fig_method} 
\end{figure*}

As motivated by the recent trend in autonomous driving~\cite{smolyanskiy2017toward, loquercio2018dronet, alexander_2019_variational_end2end}, our goal is to build a framework that can directly map the input sensory data $\mathbf{X}=(\mathcal{D}, \mathcal{P}, \mathcal{I})$, to the output steering commands $\mathbf{Y}$. To this end, we design Navigation Multimodal Fusion Network (NMFNet) with three branches to handle three visual modalities~\cite{nguyen2020autonomous}. The architecture of our network is illustrated in Fig.~\ref{Fig_method}.

In particular, we use ResNet8 to extract features from the Distance Map image and the RGB Image. The point cloud is fed into a deep network to learn learning a symmetric function on transformed elements~\cite{qi2017pointnet}.
Given the features from the point cloud branch and the RGB image branch, we first do an early fusion by concatenating the features extracted from the input cloud with the deep features extracted from the RGB image. The intuition is that since both the RGB image and the point cloud are captured using a camera with the same viewpoint, fusing their features will let the robot aware of both visual information from RGB image and geometry clue from point cloud data. This concatenated feature is then fed through two $1\times1$ convolutional layers. Finally, we combine the features from 2D-3D fusion with the extracted features from the distance map branch. The steering angle is predicted from a final fully connected layer keeping all the features from the multimodal fusion network. 

\subsubsection{Training:}
We train both networks end-to-end using the mean squared error (MSE) $L_2$ loss function between the ground-truth human actuated control, $y_i$, and the predicted control from the network $\hat{y}$:

\begin{equation}
L(y,\hat{y}) = \frac{1}{m}\sum\limits_{i = 1}^m {({y_i} - {{\hat y}_i})^2}
\end{equation}

\subsubsection{Implementation:} Our network is implemented Tensorflow framework~\cite{abadi2016tensorflow}. The optimization is done using stochastic gradient descent with the fix $0.01$ learning rate and $0.9$ momentum. The input RGB image and distance map size are ($480\times640$) and ($320\times640$), respectively, while the point cloud data are randomly sampled to $20480$ points to avoid memory overloading. The network is trained with the batch size of $8$ and it takes approximately $30$ hours to train our models on an NVIDIA 2080 GPU.
\section{Result and Deployment} \label{Sec:exp}
\subsection{Result}
\textbf{Baseline:} The results of our network is compared with the following recent works: DroNet~\cite{loquercio2018dronet}, VariationNet~\cite{amini2018variational}, and Inception-V3~\cite{szegedy2016rethinking}. All the methods are trained with the data from domain randomisation. We note that DroNet
architecture is similar to our network for normal environment, as we both use ResNet8 as the backbone. However, our network does not have the colision branch as in DroNet. The results of our NMFNet are shown under two settings: with domain randomisation (NMFNet with DR), and without using training data from domain randomisation (NMFNet without DR).

\begin{table}
\centering\ra{1.4}
\caption{RMSE Scores on the Test Set}
\label{tb_result}
\renewcommand\tabcolsep{5.5pt}


\begin{tabular}{@{}rccccc@{}}
\toprule 			&		
Input           &
House             & 
City             & 
Cave             & 
Average          \\     
\midrule
DroNet~\cite{loquercio2018dronet}				        & RGB         & 0.938 & 0.664   & 0.666   & 0.756   \\
Inception-V3~\cite{szegedy2016rethinking}			    & RGB         & 1.245 & 1.115   &1.121   & 1.16 \\
VariationNet~\cite{amini2018variational}                & RGB         & 1.510 & 1.290   & 1.507   & 1.436\\
\cline{1-6}
NMFNet without DR 	        & Fusion           & 0.482 & \textbf{0.365}   & 0.367   & 0.405 \\
NMFNet with DR 	        & Fusion           & \textbf{0.480} & \textbf{0.365}   & \textbf{0.321}   & \textbf{0.389}\\

\bottomrule
\end{tabular}
\end{table}

\textbf{Results:} Table~\ref{tb_result} summarizes the regression results using Root Mean Square Error (RMSE) of our NMFNet and other state-of-the-art methods. From the table, we can see that our NMFNet outperforms other methods by a significant margin. In particular, our NMFNet trained with domain randomisation data achieves $0.389$ RMSE which is a clear improvement over other methods using only RGB images such as DroNet~\cite{loquercio2018dronet}. This also confirms that using multi visual modalities input as in our fusion network is the key to successfully navigate in complex environments.

\begin{figure*}[h]
  \centering
  
    \subfigure[Input Image]{\includegraphics[width=0.44\linewidth, height=0.33\linewidth]{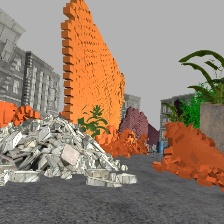}}
    \subfigure[RGB]{\includegraphics[width=0.44\linewidth, height=0.33\linewidth]{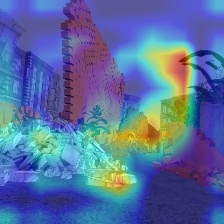}}
\subfigure[RGB + Point Cloud]{\includegraphics[width=0.44\linewidth, height=0.33\linewidth]{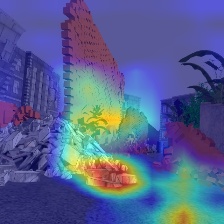}}
\subfigure[Fusion]{\includegraphics[width=0.44\linewidth, height=0.33\linewidth]{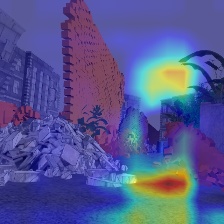}}

    \vspace{3ex}
    
 \caption{The activation map when different modalities are used to train the network. From left to right: \textbf{(a)} The input RGB image. \textbf{(b)} The activation map of the network uses only the RGB image as input. \textbf{(c)} The activation map of the network uses both RGB image and point cloud as input. \textbf{(d)} The activation map of the network uses fusion input (both RGB, point cloud and distance map). Overall, the network uses fusion input with all three modalities produces the most reliable heat map for navigation.}
 \label{Fig_activation}
\end{figure*}

\textbf{Visualization:}
As deep learning is usually considered as a black box without a clear explanation what has been learn, to further verify the contribution of each modality and the network, we employ Grad-CAM~\cite{selvaraju2017grad} to visualize the activation map of the network when different modality is used. Fig.~\ref{Fig_activation} shows the qualitative visualization under three input settings: RGB, RGB + point cloud, and fusion. From Fig.~\ref{Fig_activation}, we can see that from a same viewpoint, the network that uses fusion data makes the most logical decision since its attention lays on feasible regions for navigation, while other networks trained with only RGB image or RGB + point cloud show more noisy attention.

\begin{figure}[h] 
    \centering
    \includegraphics[width=0.99\linewidth, height=0.6\linewidth]{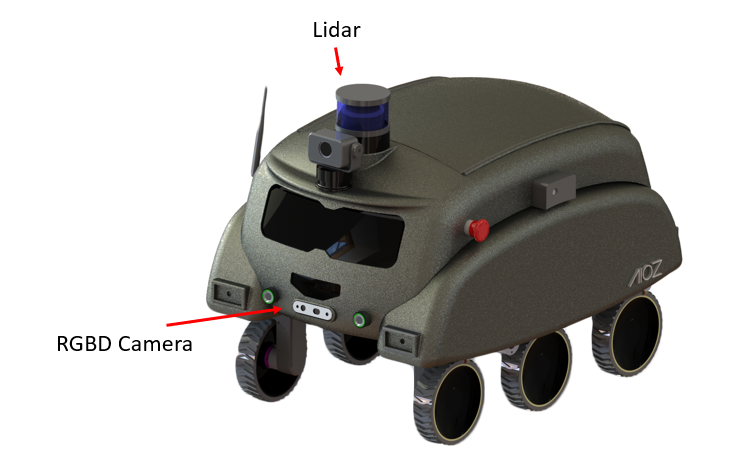}
 
    \vspace{2ex}
    \caption{An overview of BeetleBot and its camera system.}
    \label{Fig_bettlebot} 
\end{figure}

\subsection{Deployment}
\subsubsection{BeetleBot Robot:}
We deploy the trained model on BeetleBot \cite{nguyenbeetlebot}, a new mobile robot that has flexible maneuverability and strong performance to traverse through doorways, over obstacles or rough terrains that may be encountered in indoor, outdoor, or rough terrain
environments. 

BeetleBot has the novel rocker-bogie mechanism with 6-wheels.In order to go over an obstacle, the front wheels are forced against the obstacle by the back two wheels. The rotation of the front wheel then lifts the front of the vehicle up and over the obstacle. The middle wheel is then pressed against the obstacle by the rear wheels and pulled against the obstacle by the front until it is lifted up and over. Finally, the rear wheel is pulled over the obstacle by the front two wheels. During each wheel’s traversal of the obstacle, forward progress of the vehicle is slowed or completely halted. Fig. \ref{Fig_bettlebot} shows an overview of the robot.

\vspace{2cm}
 
\begin{figure}[!t] 
    \centering
    \includegraphics[width=0.99\linewidth, height=0.7\linewidth]{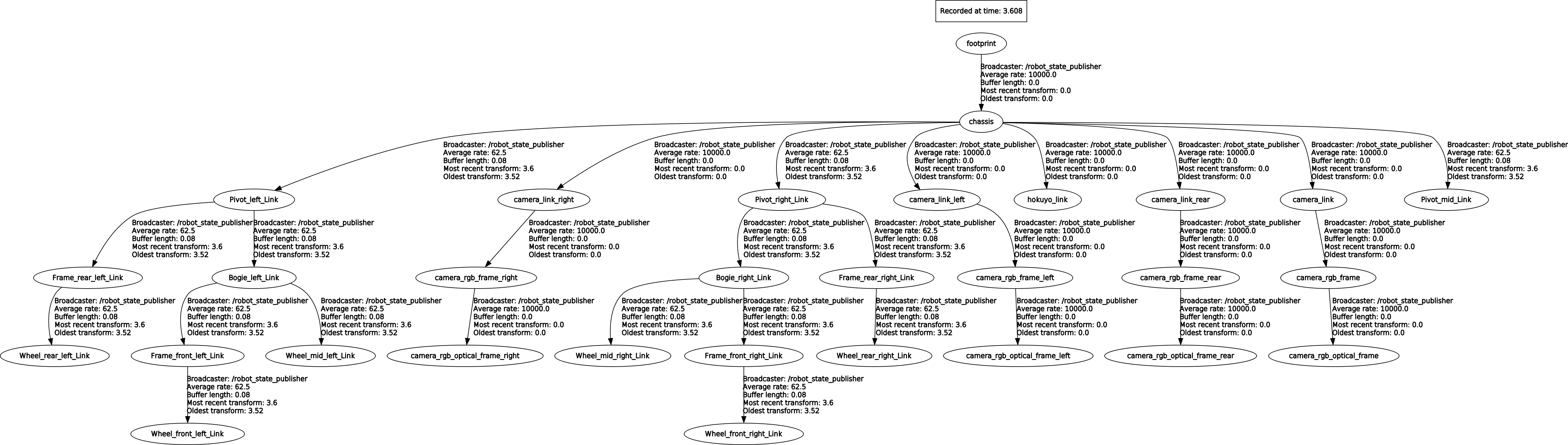}
 
    \vspace{2ex}
    \caption{The TF tree of BeetleBot.}
    \label{Fig_tf} 
\end{figure}

\begin{figure}[!t] 
    \centering
    \includegraphics[width=0.99\linewidth, height=0.6\linewidth]{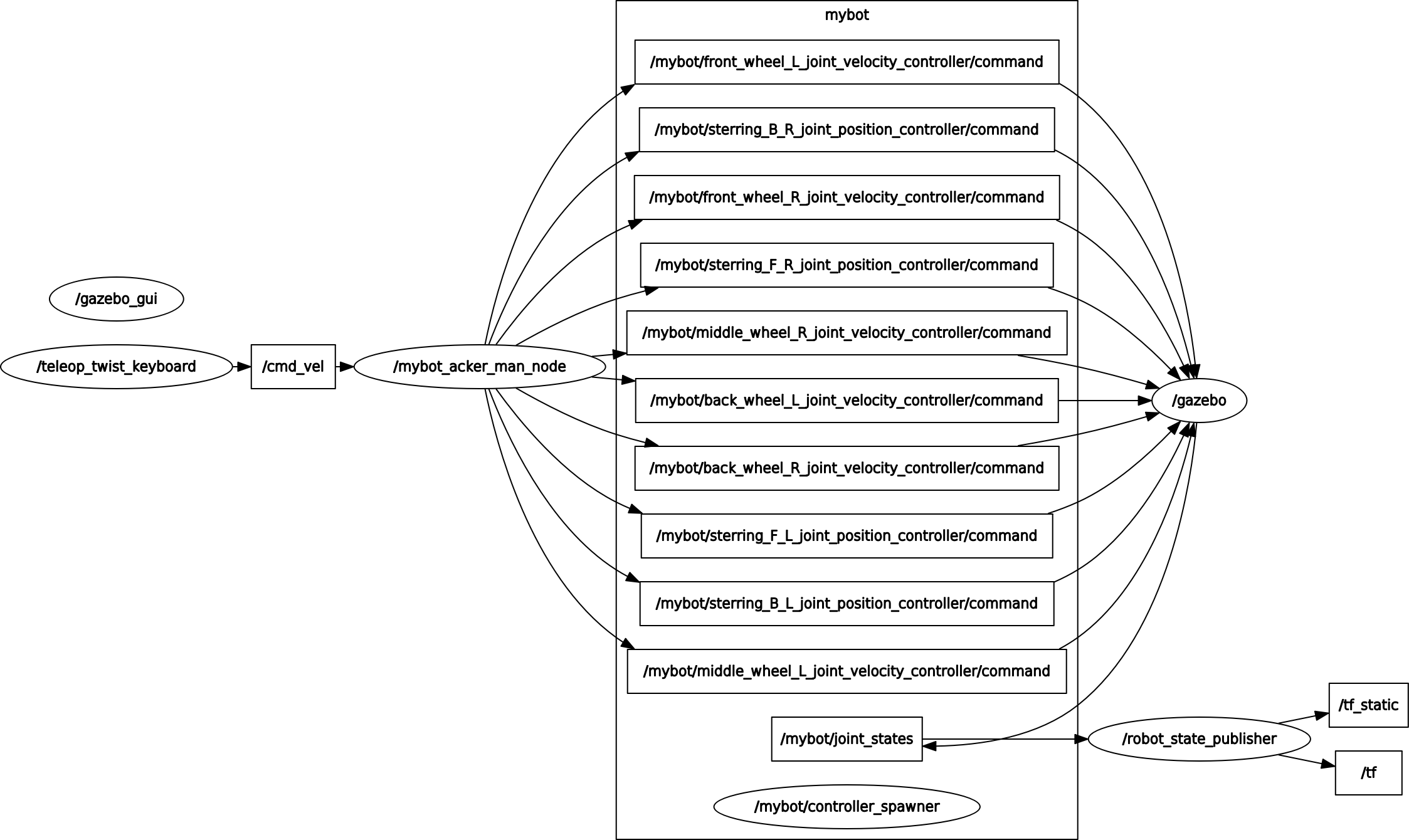}
 
    \vspace{2ex}
    \caption{The rqt\_graph of a velocity command on BeetleBot.}
    \label{Fig_cmd} 
\end{figure}

In BeetleBot, we use a RealSense camera to capture the front RGB images and point cloud, while the laser distance map is reconstructed from a RPLiDAR laser scanner. Our NMFNet runs on a Nvidia Jetson embedded board and produces velocity commands from the visual fusion input. Fig.~\ref{Fig_tf} shows the TF tree of BeetleBot in simulation, and Fig.~\ref{Fig_cmd} shows the rqt\_graph of the execution of velocity command on Gazebo.

\subsubsection{Installation:} 
Our system is tested on Ubuntu 16.04 and ROS Kinetic as well as Ubuntu 18.04 and ROS Melodic. Apart from ROS and Gazebo, the system requires the following packages:

\subsubsection{Virtual Environment:} It is recommended to install Python virtual environment for testing and deployment. The virtual environment can be installed, created, and activated via following commands:

\begin{verbatim}
- pip install virtualenv
\end{verbatim}

\begin{verbatim}
- virtualenv mynav
\end{verbatim}

\begin{verbatim}
- source mynav/bin/activate
\end{verbatim}

\subsubsection{Tensorflow:} Install Tensorflow in your Python virtual environment:

\begin{verbatim}
- pip install tensorflow-gpu
\end{verbatim}

\subsubsection{Tensorflow for Nvidia Jetson Platform:} To deploy the deep learning model on the Nvidia Jetson embedded board, we need to install ROS and Tensorflow on Nvidia Jesson. The ROS installation can be installed as usual on our laptop, while the Tensorflow installation can be done as follows:

\begin{verbatim}
- Download and install JetPack from NVIDIA-website*
\end{verbatim}

\begin{verbatim}
- pip install numpy grpcio absl-py py-cpuinfo psutil 
- pip install portpicker six mock requests gast h5py 
- pip install astor termcolor protobuf 
- pip install keras-applications keras-preprocessing 
- pip install wrapt google-pasta setuptools testresources
\end{verbatim}

\begin{verbatim}
- pip install --pre --extra-index-url NVIDA-link**
\end{verbatim}

* \url{https://developer.nvidia.com/embedded/jetpack} \\
** \url{https://developer.download.nvidia.com/compute/redist/jp/v42 tensorflow-gpu}

\subsubsection{cv-bridge:} The ROS cv-bridge package is necessary for getting data from the input camera. Install it with: 

\begin{verbatim}
- sudo apt update
\end{verbatim}

\begin{verbatim}
- sudo apt install -y ros-kinetic-cv-bridge
\end{verbatim}

\begin{verbatim}
- sudo apt install -y ros-kinetic-vision-opencv
\end{verbatim}

\subsubsection{Deployment}\label{sub_sec_results}

To deploy the trained model to the robot, we follow three main steps. We recommend the readers to our project website for detailed instructions.

\textbf{Step 1:} Getting the images from the input camera with the ROS cv-bridge package. Other input data such as laser images and point cloud should be also acquired at this step.

\textbf{Step 2:} Given the input data, doing the inference with the deep network to generate the outputted control command for the robot.

\textbf{Step 3:} The outputted control signal is formed as a Twist() ROS topic with linear speed and angular speed, then sent to the controller to move the robot accordingly.

\begin{figure*}[h]
  \centering
  
    \subfigure[Natural cave]{\includegraphics[width=0.45\linewidth, height=0.33\linewidth]{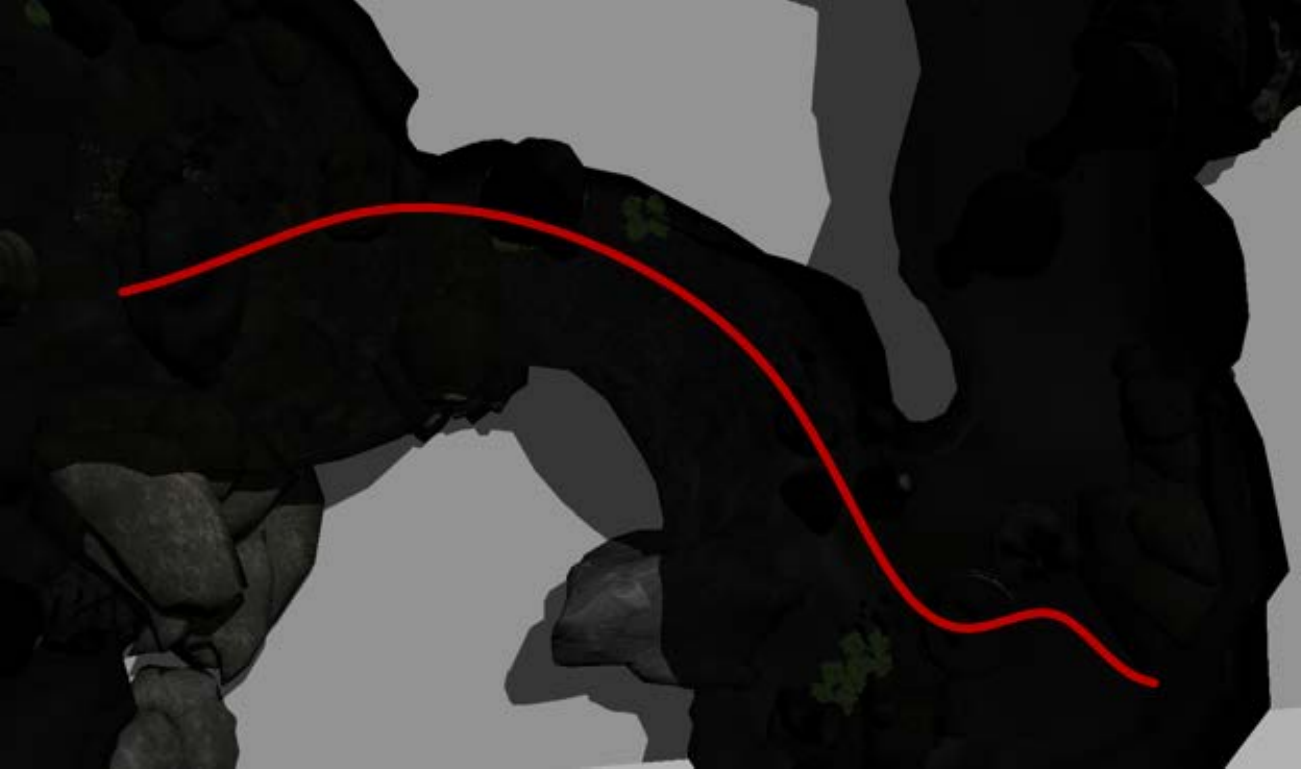}}
    \subfigure[Collapsed city]{\includegraphics[width=0.45\linewidth, height=0.33\linewidth]{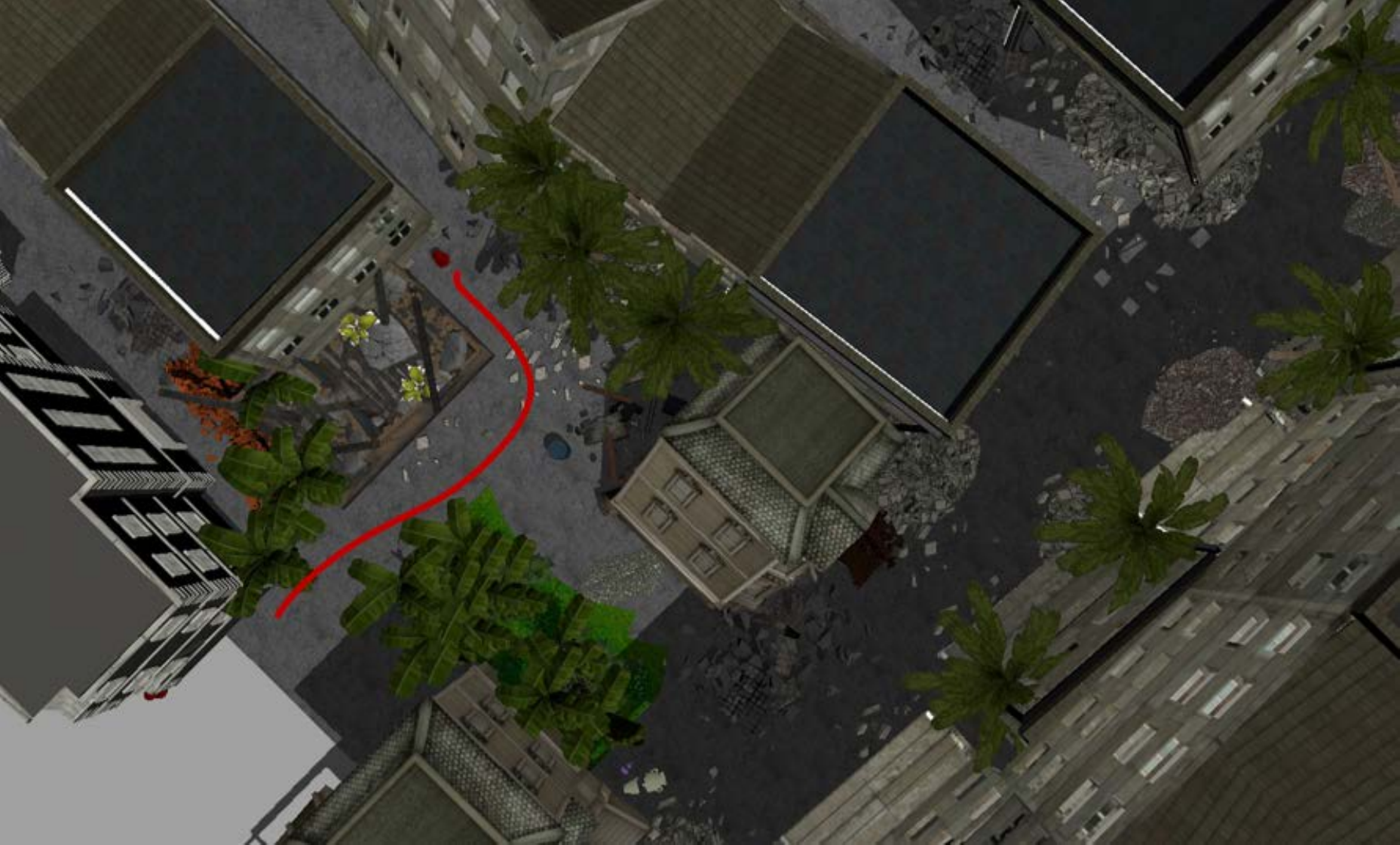}}
\subfigure[Collapsed house]{\includegraphics[width=0.45\linewidth, height=0.33\linewidth]{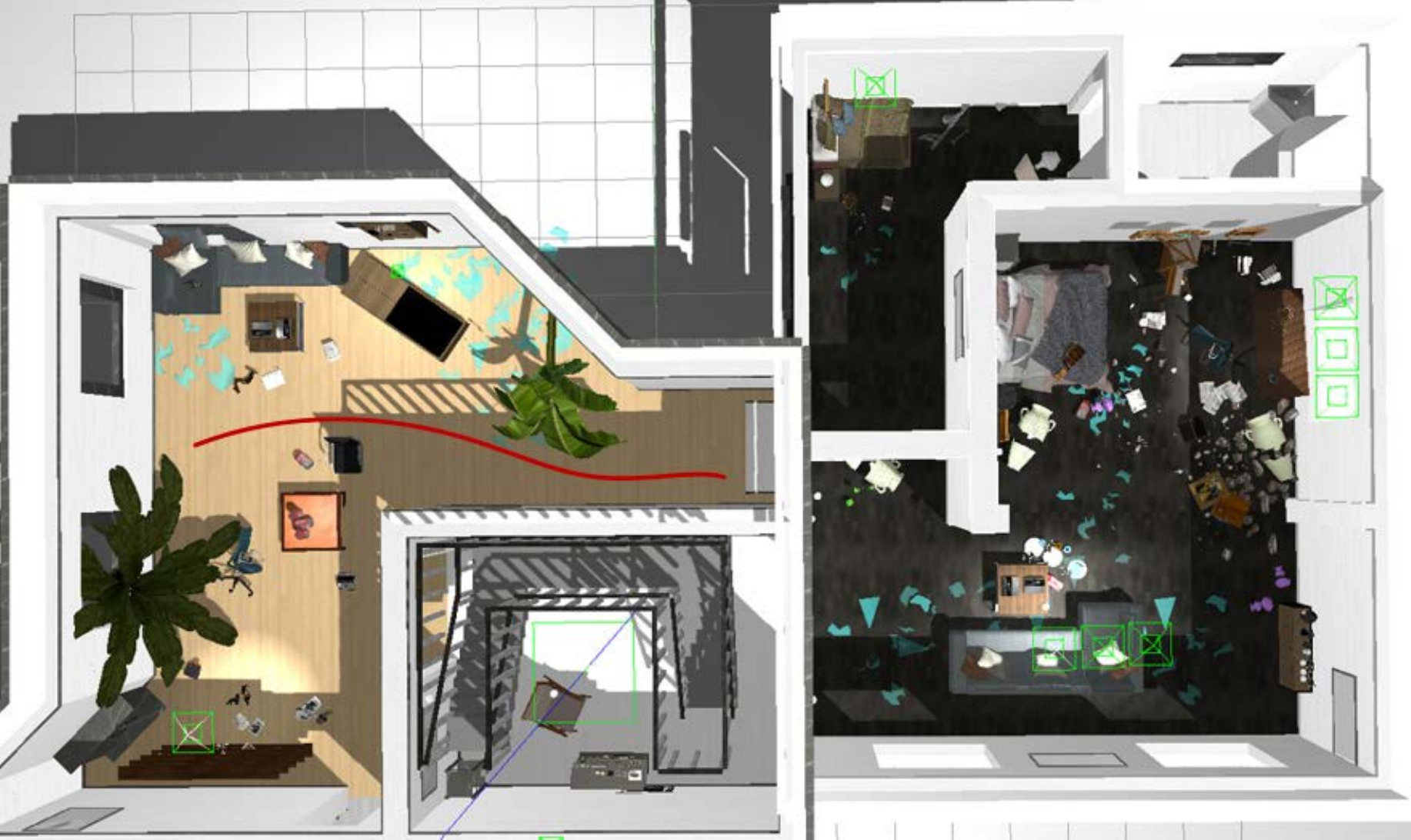}}
\subfigure[Normal house]{\includegraphics[width=0.45\linewidth, height=0.33\linewidth]{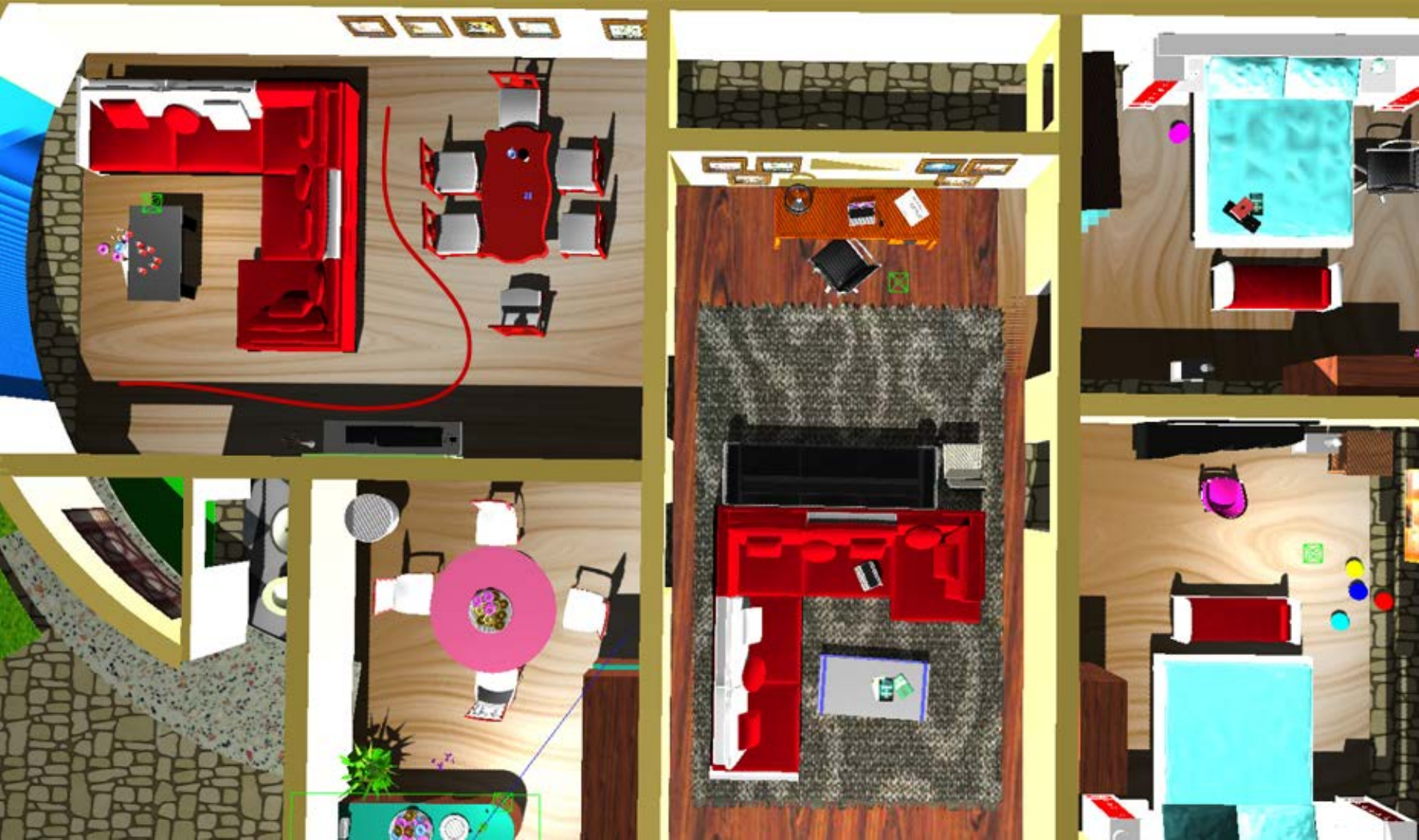}}

    \vspace{3ex}
    
 \caption{Example trajectories (denoted as red line) performed by our robot in different simulation environments.}
 \label{Fig_trajectory}
\end{figure*}

These three main steps are repeated continuously and therefore the robot will be controlled based on the learned policy during the training.

Fig. \ref{Fig_trajectory} shows some example trajectory of our mobile robot in different simulated environments. Qualitatively, we observe that the robot can navigate smoothly in the normal house or man-made road environments. In the complex environment, the distance that robot can travel is very fluctuating (i.e., ranging from $0.5m$ to $6m$) before colliding with the big obstacles. In many cases, the robot cannot go further since it cannot cross the debris or small objects on the floor. Furthermore, when we use the trained policy of the complex environment and apply it to normal environment, the results are surprisingly very promising. More qualitative results can be found in our supplemental video.

\section{Conclusions and Future Work}\label{Sec:con}

We propose different deep architectures for autonomous navigation in both normal and complex environments. To handle the complexity in challenging environments, our fusion network has three branches and effectively learns the visual fusion input data. Furthermore, we show that the use of mutilmodal sensor data is essential for autonomous navigation in complex environments. Finally, we show that the learned policy from the simulated data can be transferred to the real robot in real-world environments.

Currently, our networks show limitation on scenarios when the robot has to cross small debris or obstacles. In the future, we would like to quantitatively evaluate and address this problem. Another interesting direction is to combine our method with uncertainty estimation~\cite{richter2017safe} or a goal-driven navigation task~\cite{gao2017intention} for more wide-range of applications.


\section*{}
\addcontentsline{toc}{section}{}

\section{Authors Biographies}
\textbf{Anh Nguyen} is a Research Associate at the Hamlyn Centre, Department of Computing, Imperial College London, working on vision and learning-based methods for robotic surgery and autonomous navigation. He received his Ph.D. in Advanced Robotics from the Italian Institute of Technology (IIT) in 2019. His research interests are in the intersection between computer vision, machine learning and robotics. Dr. Anh Nguyen serves as a reviewer in numerous research conferences and journals (e.g, ICRA, IROS, RA-L). He is a member of the IEEE Robotics and Automation Society.    

\textbf{Quang D. Tran} is currently Head of AI at AIOZ, a Singapore-based startup. In 2017, he co-founded AIOZ in Singapore and led the development of “AIOZ AI” until now, which resulted with various types of AI products and high-quality research, especially in robotics and computer vision.

\bibliographystyle{class/IEEEtran}
\bibliography{class/ref}

\end{document}